\documentclass{article} 
\usepackage{iclr2017_conference,times}
\usepackage{hyperref}
\usepackage{url}

\usepackage{amssymb}
\usepackage{amsthm}
\usepackage{stfloats}
\usepackage{amsmath}
\usepackage{subcaption}

\usepackage[]{algorithm2e}
\usepackage{listings}
\usepackage{url}
\usepackage{proof}

\usepackage{hyperref}

\usepackage[pdftex]{graphicx}
\title{Linear Time Complexity Deep Fourier\\Scattering Network and Extension to\\Nonlinear Invariants}

\author{Randall Balestriero\\
Department of Electrical and Computer Engineering\\
Rice University\\
Houston, TX 77005, USA \\
\texttt{randallbalestriero@gmail.com} \\
\AND
Herv\'e Glotin \\
DYNI, LSIS, Machine Learning and Bioacoustics team\\
AMU, University of Toulon, ENSAM, CNRS\\
La Garde, France\\
\texttt{glotin@univ-tln.fr} \\
}

\begin{document}

\maketitle

\begin{abstract}In this paper we propose a scalable version of a state-of-the-art deterministic time-invariant feature extraction approach based on consecutive changes of basis and nonlinearities, namely, the scattering network. The first focus of the paper is to extend the scattering network to allow the use of higher order nonlinearities as well as extracting nonlinear and Fourier based statistics leading to the required invariants of any inherently structured input. 
In order to reach fast convolutions and to leverage the intrinsic structure of wavelets, we derive our complete model in the Fourier domain. In addition of providing fast computations, we are now able to exploit sparse matrices due to extremely high sparsity well localized in the Fourier domain. 
As a result, we are able to reach a true linear time complexity with inputs in the Fourier domain allowing fast and energy efficient solutions to machine learning tasks. Validation of the features and computational results will be presented through the use of these invariant coefficients to perform classification on audio recordings of bird songs captured in multiple different soundscapes. 
In the end, the applicability of the presented solutions to deep artificial neural networks is discussed.
\end{abstract}

\section{Introduction and Scattering Network}
\subsection{Background}
Invariants 
 are the gems of machine learning enabling key latent space representations of given inputs. Following this analogy, precious invariants shine out by being discriminative enough to detect changes in the underlying data distribution yet with bounded variations to ensure stable and robust representations. The motivation to find informative invariants as latent representations is becoming the principal focus from deep learning to signal processing communities aiming to tackle most machine learning tasks. Undoubtedly, given infinite datasets and computational power, learning invariants will lead to the fittest descriptors. However, nowadays problems do not fit this situation forcing the use of nested architectures and supercomputers to reach, in the limit, these utopian descriptors.
As an alternative, the scattering network  \citep{m1,bruna2013invariant,anden2014deep} provides a deterministic transformation of a given input signal $x$ through a cascade of linear and nonlinear operators which do not commute. The linear operator is able via a dimensional increase to linearize the representation in the sense that $x$ can be expressed as a linear combination of basic yet fundamental structures. This linear transformation is in practice a wavelet transform but can be generalized to any complete or over-complete change of basis. Originally, these wavelet transforms were used with an over-complete basis derived from Morlet and Gabor wavelets \citep{mallat1989theory}. Recently, a discrete wavelet transform scheme \citep{mallat1999wavelet} and specifically a Haar transform \citep{chen2014unsupervised} has been used instead to reduce the computational overload of the scattering transform. This framework, however, is not suited for general tasks due to poor frequency resolution of one wavelet per octave and the not continuously differentiable Haar wavelet \citep{graps1995introduction} making it unsuitable for biological and natural waveforms detection. Following the change of basis, a k-Lipschitz nonlinear operator is applied which must also be contractive to enforce space contraction and thus bound the output variations \citep{mallat2016understanding}. The nonlinearity used in the scattering network is the complex modulus which is piecewise linear. This surjection aims to map the transformation into a subspace of smaller radius $\{ | x | \mid x \in \Omega \} \subset \Omega $. As a result, one can see these successions of transforms as a suite of expansion and contraction of a deeper and deeper signal representation in the hope to decode, detect and separate all the underlying energy sources. These layered representations however still contain the time dimension and are thus overall extremely sparse and not translation invariant. This time sensitivity motivates the aggregation of the time dimension. This led to the scattering coefficients per se which are computed through the application of the operator $S$ applied on each previously computed representation and each frequency band. It is defined as an order one statistic, the arithmetic mean, over the time dimension leading to a time-invariant central tendency description of the layered representation of $x$. The resulting scattering coefficients, when used as features in machine learning tasks, led to state-of-the-art results in music genre classification \citep{chen2013music}, texture classification \citep{Sifre_2013_CVPR,bruna2013invariant} and object classification \citep{oyallon2015deep}.
In this paper, we present a modification of the standard scattering network by replacing the complex modulus nonlinearity with a quadratic nonlinearity in order to increase the SNR while allowing to compute the complete scattering network without leaving the Fourier domain, which was before necessary after each level of decomposition.

\subsection{Scattering Network}
\begin{figure}
    \centering
    \includegraphics[scale=0.32]{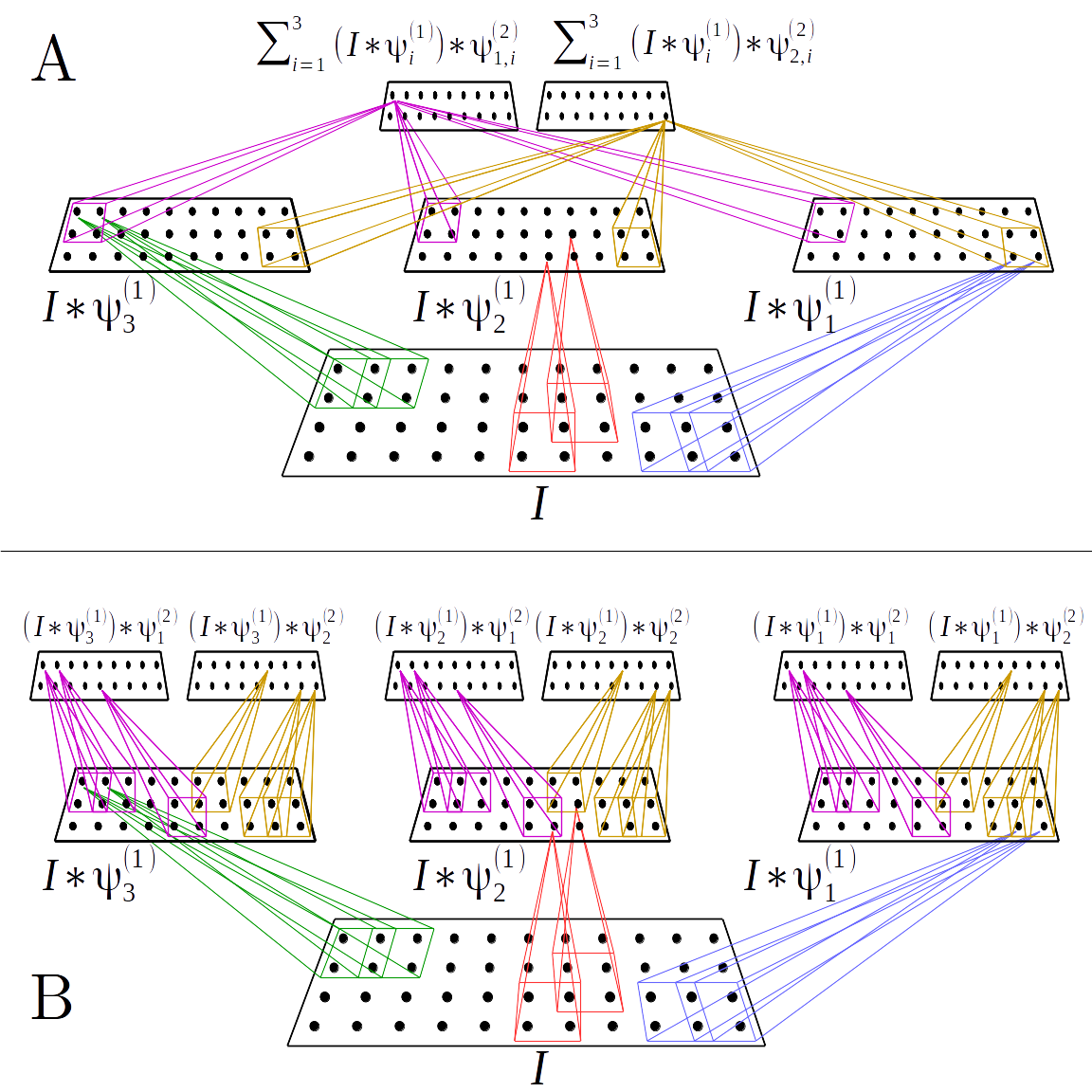}
    \caption{Architecture difference between the CNN (A) and the scattering network (B) without depiction of the computation of the scattering coefficients which are obtained after averaging over each obtained representation.}
    \label{cnn_scattering}
\end{figure}

We now present formally all the steps involved in the scattering network in order to clarify notations and concepts while making it as explicit as possible to easily develop our extensions. For readers more familiar with neural networks, it is first important to note that this framework can be seen as a restricted case of a Convolutional Neural Network \citep{lecun1995convolutional} where the filters are fixed wavelets and the nonlinearity is the complex modulus as well as some topological differences such as depicted in Fig.\ref{cnn_scattering} . The scattering coefficients are then computed through time averaging of each representation.

\subsubsection{Hierarchical Representation}
By definition a scattering network can have any fixed number of layers $L$ defined a priori. These layers are ordered in a hierarchical fashion so that the output of layer $l$ is the input of layer $l+1$. In the following, many presented properties and definitions hold for all $l \in \{1,...,L\}$.
Each layer $l$ uses a specific filter-bank made of band-pass filters $\psi^{(l)}_{\lambda}$ derived by scaling the mother wavelet of layer $l$ denoted as $\psi_0^{(l)}$ in the time domain and $\hat{\psi}_{0}^{(l)}$ in the Fourier domain. The dilation factors are denoted by the subscript $\lambda$ leading to
\begin{equation}
\psi^{(l)}_{\lambda}=\frac{1}{\lambda}\psi_0^{(l)}(\frac{t}{\lambda}) \iff \hat{\psi}^{(l)}_{\lambda}=\hat{\psi}_0^{(l)}(\lambda \omega) 
\end{equation}\label{scaling}
The collection of scaling factors $\lambda$ for layer $l$ is denoted by $\Lambda^{(l)}$. The only admissibility condition that each filter must satisfy is to have zero mean:
\begin{equation}
\int \psi^{(l)}_{\lambda}(t) dt=0 \iff \hat{\psi}^{(l)}_{\lambda}(0)=0, \forall \lambda \in \Lambda^{(l)},
\end{equation}\label{condition}
which has the following equivalent sufficient condition on each mother wavelet
\begin{equation}
\int \psi_{0}^{(l)}(t) dt =0\iff \hat{\psi}_0^{(l)}(0)=0.
\end{equation}
The finite set of continuous scale factors needed to derive the filer-bank is given as a geometric progression governed by two hyper-parameters, the number of wavelets per octave $Q$ and the number of octave to decompose $J$. The $Q$ parameter, also called quality criteria, defines the frequency resolution, the greater it is the finer the resolution is but the more redundant will be the representation. The $J$ parameter defines the number of octave to decompose. Since these parameters can be layer specific we now denote them as $Q^{(l)}$ and $J^{(l)}$. We thus have
\begin{equation}
\Lambda^{(l)}=\{2^{i/Q^{(l)}} | i= 0,...,J^{(l)}\times Q^{(l)}\}.
\end{equation} 
When the $L$ filter-banks are generated, it is possible to compute the $L$ representations by iteratively applying the filter-banks and the nonlinearity.
As a result, the $l^{th}$ representation indexed by the time dimension $t$ with the $l$ first scales as hyper-parameters is given by:
\begin{equation}
\begin{matrix}
&X^{(0)}[x](t):=x(t)\\
&X^{(1)}_{\lambda_1}[x](t):=| (X^{(0)}[x] * \psi^{(1)}_{\lambda_1})(t)|, \forall \lambda_1 \in \Lambda_1\\
&X^{(l)}_{\lambda_1,...,\lambda_l}[x](t)=|(X^{(l-1)}_{\lambda_1,...,\lambda_{l-1}}[x] * \psi^{(l)}_{\lambda_l})(t)|,\\
&\forall \lambda_1 \in \Lambda^{(1)},...,\lambda_l \in \Lambda^{(l)}
\end{matrix}
\end{equation}
One can notice that $X^{(1)}_{\lambda_1}[x]$ coefficients form the well known wavelet transform or scalogram. We now denote by $X^{(l)}[x]$ the complete time-frequency representation for layer $l$ if the scales are not necessary in the context.

Thinking of deeper layers as representations of more abstract concepts is inappropriate and thus not analogous to deep neural networks representations simply because deeper layer filters are not linear combination of the first layer filters. Since the used filters are renormalized to satisfy the Littlewood-Paley condition, the energy contained in each layer decays exponentially  \citep{waldspurger2016exponential}. As a result, deeper layer will contain less and less energy until all events have been captured and all the next layers are zeros.
With this renormalization, inverting one change of basis is instantaneous, simply add up together all the coefficients obtained from the filters application:
\begin{equation}
x(t)=\sum_{\lambda_1 \in \Lambda_1}(X^{(0)}[x] * \psi^{(1)}_{\lambda_1})(t).
\end{equation}
This last property highlights one motivation of dimensional increase, event or structure separation. It is now possible to rewrite the treated signal as a combination of fundamental structures, namely, the responses of the signal with the filters linearizing the events w.r.t the new $\lambda$ dimension.
\subsubsection{Scattering Coefficients}
For each of the $L$ representations, one can extract the scattering coefficients by applying a scaling function $\phi^{(l)}$ on $X^{(l)}[x]$ leading to the time invariant representation $S^{(l)}[x]$. The scaling function acts as a smoothing function on a given time support and satisfies
\begin{equation}
\int \phi^{(l)}(t) dt=1\iff \hat{\phi}^{(l)}(0)=1, \forall l.
\end{equation}\label{condition2}
The scaling function is usually a Gaussian filter with layer dependent standard deviations $1/\sigma^{(l)}$ in the time domain and $\sigma^{(l)}$ in the frequency domain defined as
\begin{equation}
\hat{\phi}^{(l)}(\omega)=e^{-\frac{\omega^2}{2\sigma^{(l)2}}}.
\end{equation}
The greater the standard deviation is in the physical domain the more time invariant are the scattering coefficients. Ultimately, we reach global time-invariance and the scattering operator $S[x]$ reduces to an arithmetic mean over the input support. Since only the standard deviation of the scaling function is layer dependent, we denote by $\phi^{(l)}$ the $l^{th}$ scaling function generated using $\sigma^{(l)}$. We can thus define the scattering operators as
\begin{equation}
\begin{matrix}
S^{(0)}[x](t):=(X^{(0)}[x]*\phi^{(0)})(t),\\
S^{(1)}_{\lambda_1}[x](t):=(X^{(1)}_{\lambda_1}[x]*\phi^{(1)})(t), \forall \lambda_1 \in \Lambda^{(1)}\\
S^{(l)}_{\lambda_1,...,\lambda_l}[x](t):=(X^{(l)}_{\lambda_1,...,\lambda_l}[x]* \phi^{(l)})(t)\\
\forall \lambda_1 \in \Lambda^{(1)},...,\lambda_l \in \Lambda^{(l)}
\end{matrix}
\end{equation}
For machine learning tasks, using global time-invariance yield robust yet biased time invariant descriptors due to too much many-to-one possible mappings. As a result, a local or windowed scattering transform has been used \citep{bruna2013invariant} leading to only local time-invariance through a smaller time support for the scaling function. This tweak is possible in computer vision tasks where each input is of the same size and the role of $\phi$ is to bring local diffeomorphism invariance which has been shown to smooth the underlying manifold \citep{zoran2011learning,zoran2012natural}. However, for audio tasks and more general problems, this constant input size is rare forcing the use of $\phi$ for aggregating the time dimension. In fact, local smoothing would thus not be feasible since it would lead to different feature vectors sizes for different input sizes being of great difficulty to deal with. This is why we resolve to maintain global time invariant descriptors as well as developing new nonlinear operators in order to enrich the invariant input description and reduce the bias of our new feature vector compared to the scattering coefficients.

\section{Extensions}
\subsection{Higher Order Nonlinearity}\label{nonlinearity_official}
The usual nonlinearity applied in a scattering network is the complex modulus. This nonlinearity is not everywhere differentiable but is contractive leading to an exponential decay in the energy distribution over the layers. However, as pointed out in \citep{waldspurger2016exponential}, higher order nonlinearity might be beneficial sparsity-wise and to increase the SNR.
As a result, we chose to use a continuously differentiable second order nonlinearity which will have the beneficial property of adapting its contractive property for irrelevant inputs while maintaining bounded variations. This nonlinearity is defined as
\begin{equation}
\mathcal{P}[c]=|c|^2,\; \forall c \in \mathbb{C}.
\end{equation}
\begin{proof}
We now prove the adaptive k-Lipschitz property of our nonlinearity
\begin{align*}
||\mathcal{P}[a]-\mathcal{P}[b]||=&|| |a|^2-|b|^2 ||\\
=&|| (|a|-|b|)(|a|+|b|) || \\
=&(|a|+|b|)|| |a| - |b| || \\
\leq & (|a| + |b|)||a-b|| \\
=& K(a,b) ||a-b||
\end{align*}
\end{proof}
Since the input signal is renormalized so that $||x||_1=1$, we have that $|a| +|b| \in [0,1[$. As a result, given the inputs constraints, $\mathcal{P}$ is a contractive operator with bounded variations. Yet, the degree of contraction will vary given the input amplitudes leading to a better SNR.
This means that high amplitudes resulting from close match between the filter and the signal will be efficiently represented whereas small amplitude coefficients resulting from noise filtering and mismatches between the filter and the signal will be highly contracted. Since in practice wavelet filters catch the relevant events, this property allows high quality representations.
This change will not only increase the relative sparsity of the representations but also allow us to perform some major computational tricks as describe in the following section.
As a result we define the new representations as
\begin{equation}
\begin{matrix}
&\mathcal{X}_0[x](t):=x(t)\\
&\mathcal{X}^{(1)}_{\lambda_1}[x](t):=\mathcal{P}\left[(\mathcal{X}^{(0)}[x] * \psi^{(1)}_{\lambda_1})(t)\right], \forall \lambda_1 \in \Lambda^{(1)}\\
&\mathcal{X}^{(l)}_{\lambda_1,...,\lambda_l}[x](t)=\mathcal{P}\left[(\mathcal{X}^{(l-1)}_{\lambda_1,...,\lambda_l}[x]* \psi^{(l)}_{\lambda_l})(t)\right],\\
& \forall \lambda_1 \in \Lambda^{(1)},...,\lambda_l \in \Lambda^{(l)}
\end{matrix}
\end{equation}
as well as the new scattering coefficients as
\begin{equation}
\begin{matrix}
\mathcal{S}^{(0)}[x](t):=(\mathcal{X}^{(0)}[x]* \phi^{(0)})(t),\\
\mathcal{S}^{(1)}_{\lambda_1}[x](t):=(\mathcal{X}^{(1)}_{\lambda_1}[x] * \phi^{(1)})(t), \forall \lambda_1 \in \Lambda^{(1)}\\
\mathcal{S}^{(l)}_{\lambda_1,...,\lambda_l}[x](t):=(\mathcal{X}^{(l)}_{\lambda_1,...,\lambda_l}[x] * \phi^{(l)})(t),\\
\forall \lambda_1 \in \Lambda^{(1)},...,\lambda_l \in \Lambda^{(l)}
\end{matrix}
\end{equation}

\subsection{Invariant Dispersion Coefficients}
The scattering coefficients used to characterize the signal of interest are known to be efficient for stationary inputs but not descriptive enough otherwise. We thus propose to generate complementary invariant coefficients based on a dispersion measure, the variance. As a result, these complementary coefficients derived from the second order moment will help to characterize the input leading to more discriminative features while maintaining global time invariance. 
We now define these invariant dispersion coefficients as
\begin{equation}
\begin{matrix}
\mathcal{V}^{(0)}[x] :=||\mathcal{X}^{(0)}[x]-\mathcal{S}^{(0)}[x]||^2_2,\\
\mathcal{V}^{(1)}_{\lambda_1}[x]:=||\mathcal{X}^{(1)}_{\lambda_1}[x] - \mathcal{S}_{\lambda_1}^{(1)}[x]||^2_2, \forall \lambda_1 \in \Lambda^{(1)} \\
\mathcal{V}^{(l)}_{\lambda_1,...,\lambda_l}[x]:=||\mathcal{X}^{(l)}_{\lambda_1,...,\lambda_l}[x]- \mathcal{S}^{(l)}_{\lambda_1,...,\lambda_l}[x]||_2^2.\\
\forall \lambda_1 \in \Lambda^{(1)} ,...,\lambda_l \in \Lambda^{(l)}.
\end{matrix}
\end{equation}
The resulting $\mathcal{V}^{(l)}[x]$ coefficients are thus globally time invariant whatever scaling function was used to compute $\mathcal{S}^{(l)}[x]$. In fact, these invariant dispersion coefficients represent the variance between $\mathcal{X}^{(l)}[x]$ and $\mathcal{S}^{(l)}[x]$ representations whether $\mathcal{S}^{(l)}[x]$ was globally time invariant or a smoothed version of $\mathcal{X}^{(l)}[x]$.
In order for $\mathcal{V}^{(l)}[x]$ to be invariant to random permutations as well, $S^{(l)}[x]$ should be globally translation invariant and thus also globally invariant to random permutations.
In addition, regarding the discriminative ability gained through the use of these second order statistic, we have that
\begin{equation}
\begin{matrix}
card\left(\{ s\in \mathbb{L}^2(\mathbb{C}) | \textbf{S}=(k_1,...,k_n) \}\right) \geq\\
card\left(\{ s\in \mathbb{L}^2(\mathbb{C}) | \textbf{S}=(k_1,...,k_n) \text{ and } \textbf{V}=(p_1,...,p_n)  \}\right),
\end{matrix}
\end{equation} 
where $\textbf{S}$ represents a realization of the scattering coefficients for all layers, all frequency bands, and $\textbf{V}$ a realization of the dispersion coefficients again for all layers and all frequency bands.
From this, it follows that the set of invariant coefficients $\left( \textbf{S}[x],\textbf{V}[x] \right)$ is more discriminative leading to more precise data description than when using $\left( \textbf{S}[x]\right)$ only.
The development of these presented invariant dispersion coefficients opened the door to the development of uncountably many new invariant coefficients. We now present the elaboration of the scheme in the Fourier domain and the computational tricks involved in order to reach linear complexity.
\section{Fast Implementation, Fourier Domain and Linear Complexity}
\subsection{Introduction and Sparse Storage}
One of the great benefits of the wavelet transform is the induced sparsity in the representation \citep{elad2006image,starck2010sparse}. This benefit has been well exploited when dealing with compression problems and algorithms such as the JPEG-2000 standard \citep{skodras2001jpeg} which uses wavelet transforms. For more general application however, the induced sparsity has never been leveraged and is only pointed out as a quality criteria of the representation \citep{coifman1992wavelet}. 
To emphasize this point, let consider the application of one filter. In the time domain, this is done by convolving the input with the filter. This corresponds to computing multiple local inner products between the filters and subparts of the signal. Thus it is clear that one can not know a priori which subparts of the signal are orthogonal to the filter corresponding to the positions of the zeros coefficients which are indeed unpredictable. On the contrary, applying the filter in the Fourier domain reduces  to an Hadamart product and thus the resulting support is deduced from the filter support. As a result, it is now possible to know a priori most of the zero coefficients positions since in the Fourier domain the filter is well localized.
In other words, in the time domain, one should compute the whole representation and then retrieve the position and values of the coefficients lying above a specific threshold as if it was part of a compression algorithm. In the Fourier domain, the wavelet support is a convex compact which can be computed a priori given the scale parameter and the mother wavelet and thus the nonzeros coefficients positions are known a priori independently of the input. This motivates our choice to perform our framework including the wavelet transform, the nonlinearity $\mathcal{P}$ and the invariant features extraction in the Fourier domain leading to linear complexity. Furthermore, using the Fourier domain allows us to efficiently leverage sparse matrices leading to efficient storage and memory management on energy efficient platforms such as presented in \citep{esser2015backpropagation}.
We will first present the computation of the filters in the Fourier domain as well as their convex compact support derivation.
From that we present the sparse application of the filters and how to compute the nonlinearity in Fourier. Finally, we will see that extracting the invariant features can be done efficiently leading to our main result which is a linear complexity overall framework. Concerning the Fourier transform, the Danielson-Lanczos lemma \citep{flannery1992numerical} will be used in order to provide a true $O(N \log (N))$ complexity for an input of size $N$ which is a power of $2$. As we will see, this requirement will always be fulfilled without any additional cost.
\subsection{Sparse Fourier Filters}\label{sparse_filter_implementation}
\begin{figure}[t!]
\centering
\includegraphics[width=4.5in]{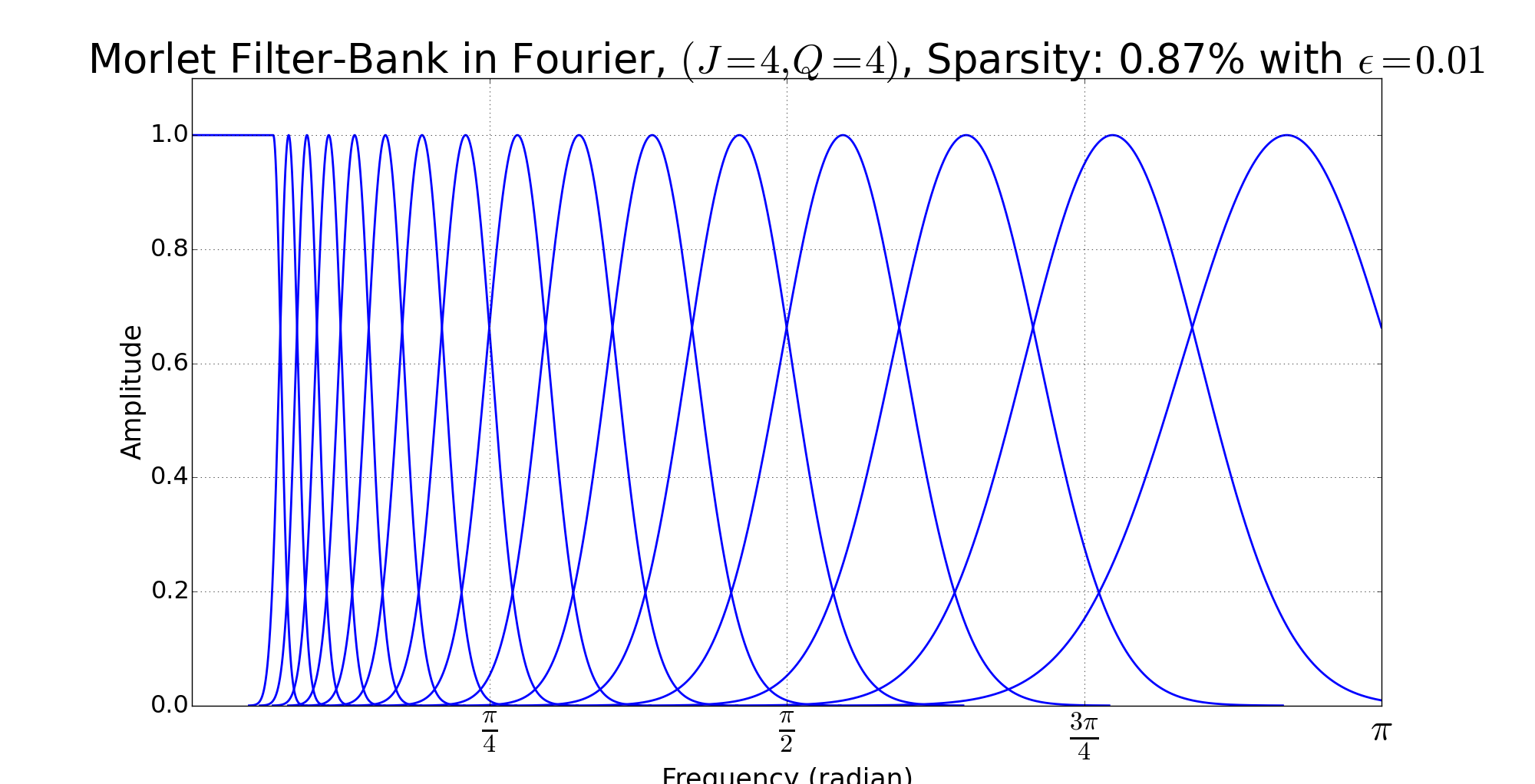}
\caption{Filter-Bank in Fourier with $4$ wavelets per octave on $4$ octaves. The sparsity is of $0.94\%$ with $\epsilon=0.01$}
\label{filters}
\end{figure}

One particularity of the continuous wavelets such as DoG, Morlet wavelets reside in their localized compact support in the Fourier domain. In our description the used wavelet will be a Morlet wavelet but this analysis can be extended to any continuous wavelet with analytical form.
We define the support of the filter $\psi^{(l)}_\lambda$ given the threshold $\epsilon$ as 
\begin{equation}
supp_\epsilon[\psi^{(l)}_\lambda]:=\{ \omega | \psi^{(l)}_\lambda (\omega) > \epsilon, \omega \in [0,2\pi] \}
\end{equation}
As presented in \citep{balestriero2015scattering} the scales define entirely the support of each wavelet. In order to develop synergistic computational tricks, we derive our framework in the Fourier domain. Let define the Morlet wavelet as
\begin{equation}
\hat{\psi}_{\mu_0,\sigma_0}(\omega)=H(\omega)e^{-\frac{(\omega-\mu_0)^2}{2\sigma_0^2}},
\end{equation}
where the parameters $\mu_0$ and $\sigma_0$ represent respectively the center frequency and bandwidth of the mother wavelet and $H$ is the step-wise function. The ratio between these two quantities will remain the same among all the filters, in fact, wavelets have a constant ratio of bandwidth to center frequency. 
These two mother hyper-parameters are taken as
\begin{equation}
\begin{matrix}
\mu_0=\frac{\pi}{2}(2^{-1/Q}+1)\\
\sigma_0=\sqrt{3}(1-2^{-1/Q}).
\end{matrix}
\end{equation}
Yet, instead of using the definition of scaling as defined in section \ref{scaling} we will use these two parameters as follows
\begin{equation}
\hat{\psi}_{\mu_0,\sigma_0}(\lambda \omega)=\hat{\psi}_{\frac{\mu_0}{\lambda},\frac{\sigma_0}{\lambda}}(\omega):=\hat{\psi}_\lambda(\omega).
\end{equation}
In fact, we have the following relation between the scale and the mother hyper-parameters
\begin{align*}
\hat{\psi}_{\mu_0,\sigma_0}(\lambda \omega)=&H(\lambda \omega )e^{-\frac{(\lambda \omega - \mu_0)^2}{2\sigma_0^2}}\\
=&H(\omega)e^{-\frac{\lambda^2(\omega - \frac{\mu_0}{\lambda})^2}{2\sigma_0^2}}\\
=&H(\omega)e^{-\frac{(\omega - \frac{\mu_0}{\lambda})^2}{2(\frac{\sigma_0}{\lambda})^2}}.\\
=&\hat{\psi}_{\frac{\mu_0}{\lambda},\frac{\sigma_0}{\lambda}}(\omega)
\end{align*}
Given this, we can compute explicitly the support of every filter $\psi^{(l)}_\lambda$.
As one can notice these filters have a convex compact support around their center frequencies:
\begin{equation}
\begin{matrix}
supp_\epsilon[\psi^{(l)}_\lambda]=\left[ \frac{\mu_0}{\lambda}-\frac{\sigma_0}{\lambda}\sqrt{-2 \log(\epsilon)}, \right. \\
\;\;\;\;\;\left.
\frac{\mu_0}{\lambda}+\frac{\sigma_0}{\lambda}\sqrt{-2 \log(\epsilon)} \right].
\end{matrix}
\end{equation}
In Fig. \ref{filters} one can see a filter-bank example where all the filters are presented in one plot demonstrating the important sparsity inherent to wavelets. Varying the $\epsilon$ parameter affects directly the number of nonzeros coefficients and we thus also present in Table \ref{table1} the exact sparsity with different input sizes and $\epsilon$ parameter.
Since the support is known a priori, it is straightforward to optimize their computation and allocation through sparse vectors leading to fast filter-bank generation as presented in Table \ref{table2} with large input sizes. In fact, the input length defines the size of the generated filter since we now perform the convolution in the Fourier domain.

\begin{table}[t!]
\centering
\caption{Sparsity in percentage of the Morlet Filter-Bank in Fourier for different signal sizes and realistic parameters $(J=9,Q=16)$.}
\begin{tabular}{|c|c|c|c|}
\hline
          & $\epsilon=0.0001$ & $\epsilon=0.0000001$ \\ \hline
$N=524288 \;\; (2^{19})$ & $98.39883\%$ & $97.89803\%$ \\ \hline
$N=1048576 \;\;(2^{20})$     & $98.39893\%$ & $97.89812\%$ \\ \hline
$N=2097152 \;\;(2^{21})$     & $98.39898\%$ & $97.89816\%$ \\ \hline
\end{tabular}
\label{table1}
\end{table}

\begin{table}[t!]
\centering
\caption{Time (in sec.) needed to compute the filter-bank given the signal size with standard parameters $J=5,Q=16$ growing linearly with the input size on $1$ CPU @$1.4$GHz.}
\begin{tabular}{|c|c|c|c|c|c|c|c|}
\hline
Signal Size & $2^{16}$&  $2^{18}$&  $2^{20}$& $2^{22}$\\\hline
Time in sec. &$0.006$&$ 0.018 $& $ 0.066$&$ 0.262$ \\\hline
\end{tabular}
\label{table2}
\end{table}

Concerning the $\phi^{(l)}$ filter, given its bandwidth $\sigma^{(l)}$, its support is given by:
\begin{equation}
supp_\epsilon [\phi^{(l)}]=\left[-\sigma^{(l)}\sqrt{-2 \log(\epsilon)},\sigma^{(l)}\sqrt{-2 \log(\epsilon)} \right].
\end{equation}
Some examples are shown in Fig. \ref{phis} where different $\sigma^{(l)}$ are selected representing different time supports.
For each of the filters and each of the layers, the application is done through element-wise multiplication in the Fourier domain as explained in the next section where the nonlinearity will be defined.
\subsection{Nonlinearity and Filtering in Fourier}
The nonlinearity used in this framework defined in section  \ref{nonlinearity_official} is efficiently done in the Fourier domain through the following property
\begin{equation}
\mathcal{F}[|x|^2]=\mathcal{F}[x] *\mathcal{F}[x]^* 
\end{equation}
If done directly, this operation would be slower in the Fourier domain since we jump from a linear complexity to a quadratic complexity. However, one should notice from section  \ref{sparse_filter_implementation} that we are dealing with $\mathcal{F}[x]$ which are extremely sparse but most importantly with  convex compact support of size $M <<N$. Exploiting this sparsity could lead to a faster convolution which would still be of quadratic complexity w.r.t the support size. However, using the convolution theorem it is possible to perform this convolution in $M \log (M)$ complexity by applying again a Fourier transform now only on the convex compact support of $\mathcal{F}[x]$.
In order to have proper boundary condition and not the periodic assumption of the Fourier transform we use a zero-padded version of size$2M$ instead of $M$ leading to exact computation of the convolution. In addition, the support size of $2M$ is the minimum required size. For a fast Fourier transform algorithm, this has to be a power of $2$. As a result, in practice, the zero padding is done to reach the size which is the smallest power of $2$ greater than $2M$ defined as
\begin{equation}
2^{\left \lceil \log_2(2M)\right \rceil},
\end{equation}
where $\left \lceil \log_2(2M)\right \rceil$ denotes the smallest of the greater integers.
As a result in the Fourier domain we will apply another Fourier transform in order to compute this auto-correlation which will correspond to the desired nonlinearity in the time domain.
\begin{equation}
\mathcal{F}[|x|^2]=\mathcal{F}^{-1}\left[ \mathcal{F}[\mathcal{F}[x]] \bigodot  \mathcal{F}[\mathcal{F}[x]]^* \right],
\end{equation}
where $\bigodot$ is the Hadamart product, $\mathcal{F}$ is the Discrete Fourier Transform and $\mathcal{F}^{-1}$ its inverse operator.
In addition of the second Fourier transform being applied on a really small support, it is also important to note that after application of the nonlinearity the output is conjugate symmetric in the Fourier domain but since the filter-banks are always applied on $[0,\pi] $ we can store only this part for further computation and re-generate the conjugate symmetric part when applying $\phi^{(l)}$. We present this operation in the Fourier domain in Fig. \ref{example_nonlinearity}.

\begin{figure}[t!]
\centering
\includegraphics[width=5.5in]{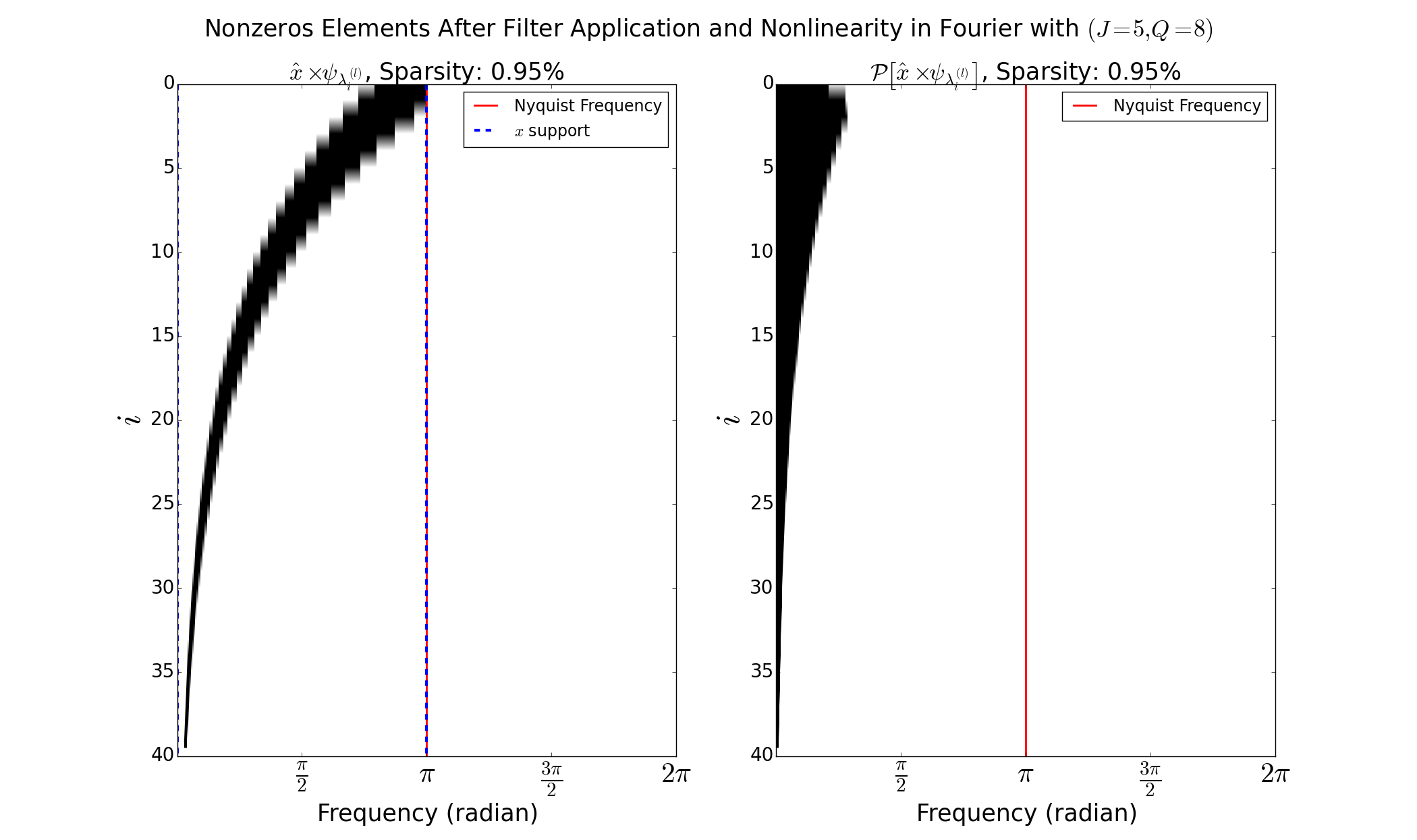}
\caption{Nonzeros elements present after application of the filters and the nonlinearity operator $\mathcal{P}$ on this representation for an input without sparsity.}
\label{fig::sparsity_matrix1}
\end{figure}

In order to highlight the high sparsity encountered in the Fourier domain when dealing with this filter application, we present in Fig. \ref{fig::sparsity_matrix1} an example where the nonzeros elements are shown. This corresponds to the first representation namely $\mathcal{X}^{(1)}[x]$. For the second representation, the input support will not be over the whole $[0,2\pi[$ domain but around $0$ and thus implies increased sparsity as demonstrated in Fig. \ref{fig::sparsity_matrix2}.
Given these two descriptions, one is able to compute $\mathcal{X}^{(l)}[x]$ for any $l$. We thus now present how to compute the scattering and dispersion coefficients given this representations in the Fourier domain.

\subsection{Scattering Coefficients Extraction}
The scattering coefficients $\mathcal{S}^{(l)}[x]$ result from the application of a Gaussian filter parameterized by its standard deviation. In the general case where global time invariance is required, this standard deviation is taken to be infinite in the time domain resulting in
\begin{equation}
\mathcal{S}^{(l)}_{\lambda_1, ... ,\lambda_l}[x]=\frac{1}{N}\sum_{t=1}^N \mathcal{X}_{\lambda_1, ... ,\lambda_l}^{(l)}[x](t).
\end{equation}
In this case, the corresponding result in Fourier is given by 
\begin{equation}
\mathcal{S}^{(l)}[x]= \mathcal{F} \left[ \mathcal{X}^{(l)}[x]\right] (\omega=0).
\end{equation}
The invariant dispersion coefficients are extracted from the Fourier transform in a straightforward manner as shown in the Appendix which results in 
\begin{equation}
||\mathcal{X}^{(l)}[x]-\mathcal{S}^{(l)}[x]||_2^2\\
=||\mathcal{F}\left[ \mathcal{X}^{(l)}[x] \right] \bigodot (1 - \mathcal{F} \left[ \phi^{(l)} \right]) ||_2^2.
\end{equation}
Thus $(1 - \mathcal{F} \left[ \phi^{(l)} \right] )$ acts as a mask to reduce the norm computation by the amount of energy captured through the scaling function application. For the case where we have global time invariance or infinite standard deviation, this mask reduces to
\begin{equation}
(1 - \mathcal{F} \left[ \phi^{(l)} \right] )(\omega)=\delta(\omega)
\end{equation}
where $\delta$ denotes the Dirac function. As a result, the dispersion coefficients can be calculated as
\begin{equation}
\mathcal{V}^{(l)}[x]=2\sum_{\omega=1/N}^{\pi }\left(\mathcal{F}\left[\mathcal{X}^{(l)}[x] \right](\omega)\right)^2,
\end{equation}
which is the L2 norm computed without taking into account the coefficient at $\omega=0$ exploiting the conjugate symmetry structure for the real input signal $x$. Conceptually, the $\mathcal{V}$ coefficients capture the remaining energy and thus ensures that for any depth of the scattering network, all the energy of the input is contained inside the computed invariants. In fact, one can see that $\mathcal{V}^{(l)}[x]=\sum_{i=l+1}^{\infty}\mathcal{S}^{(i)}[x]$.

\subsection{Scalability}
\begin{figure}[t!]
\centering
\begin{subfigure}[b]{0.45\textwidth}
    \includegraphics[width=3in]{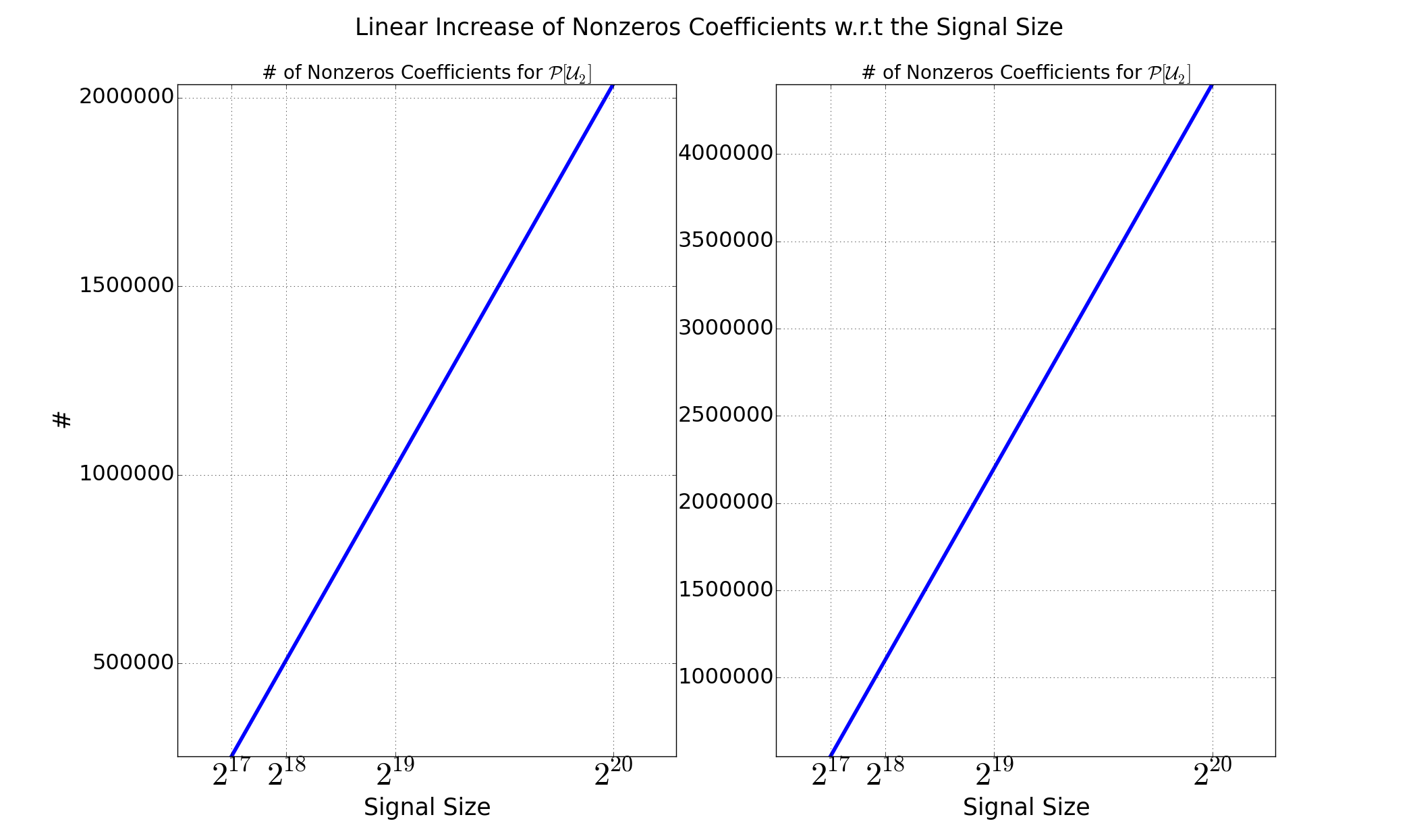}
\end{subfigure}
~
\begin{subfigure}[b]{0.45\textwidth}
    \includegraphics[width=3in]{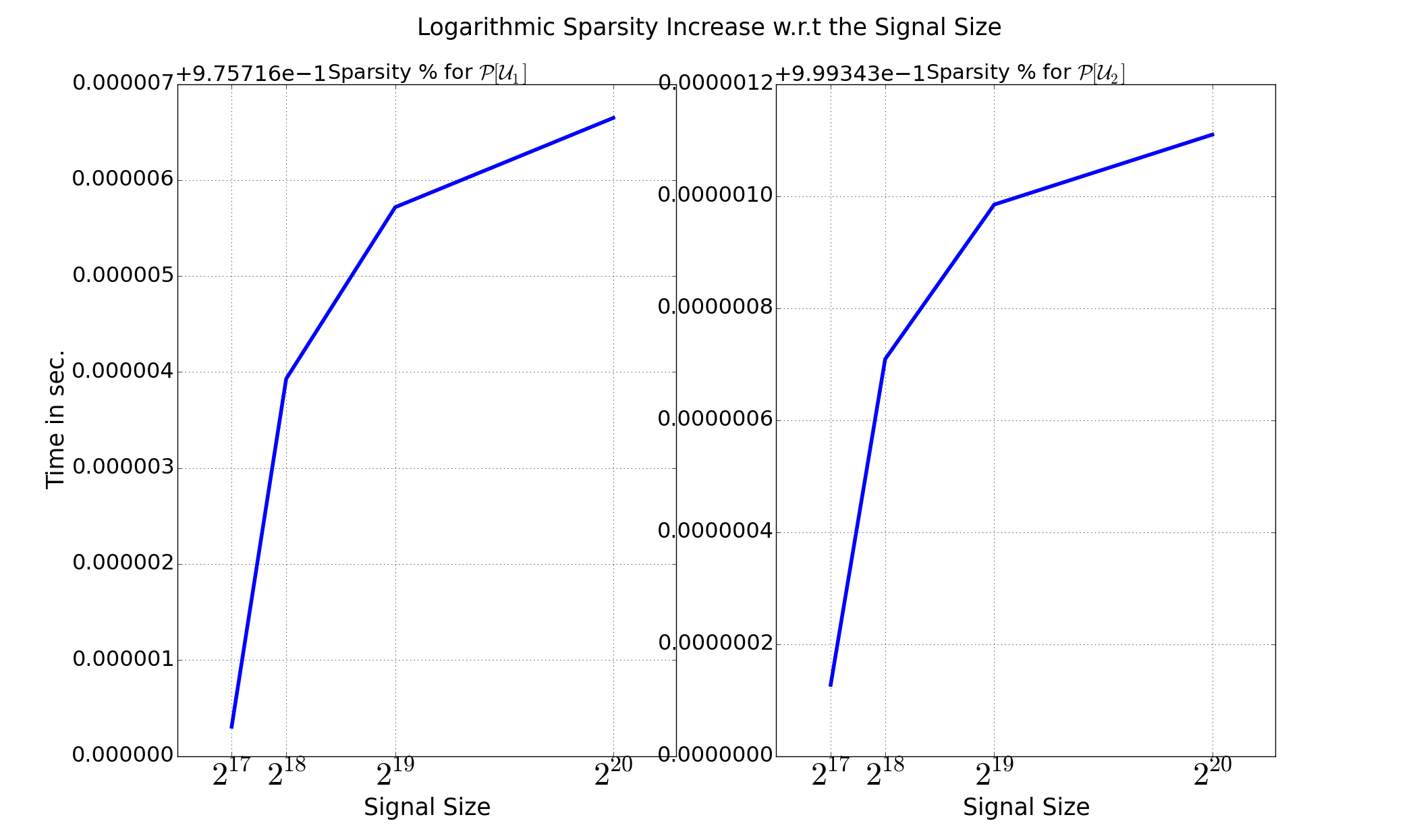}
\end{subfigure}
\caption{Left: Figure showing the increase in nonzeros coefficients is linear with respect to the input size. Right: Figure showing the increase in sparsity in our representations for the two layers. The increase is logarithmic with respect to the input size.}
\label{fig::nonzeros_increase1}
\end{figure}

We now present some results and figures in order to highlight the high scalability of our framework with respect to the input size. Note that the current implementation is done in Python. Implementing this toolbox in C is a future work which will lead to even better results than the ones displayed below which are nevertheless already astonishing. First of all, one can see in Fig. \ref{fig::nonzeros_increase1} that the number of nonzeros coefficients increase linearly with the input size. This result is important in nowadays paradigm where technologies allow extreme frequency sampling and thus input signals with high dimensions yet we aim to save as much memory and storage as possible. If we put this nonzeros coefficients in perspective with the total possible number of coefficients we obtain our sparsity coefficient which grows logarithmically with the input size as shown in Fig. \ref{fig::nonzeros_increase1}. This result shows the advantage of using sparse matrices which increases as the input size increases. The sparsity is thus in our case a justification to exploit the Fourier domain.

\begin{figure}[t!]
\centering
\includegraphics[width=5.5in]{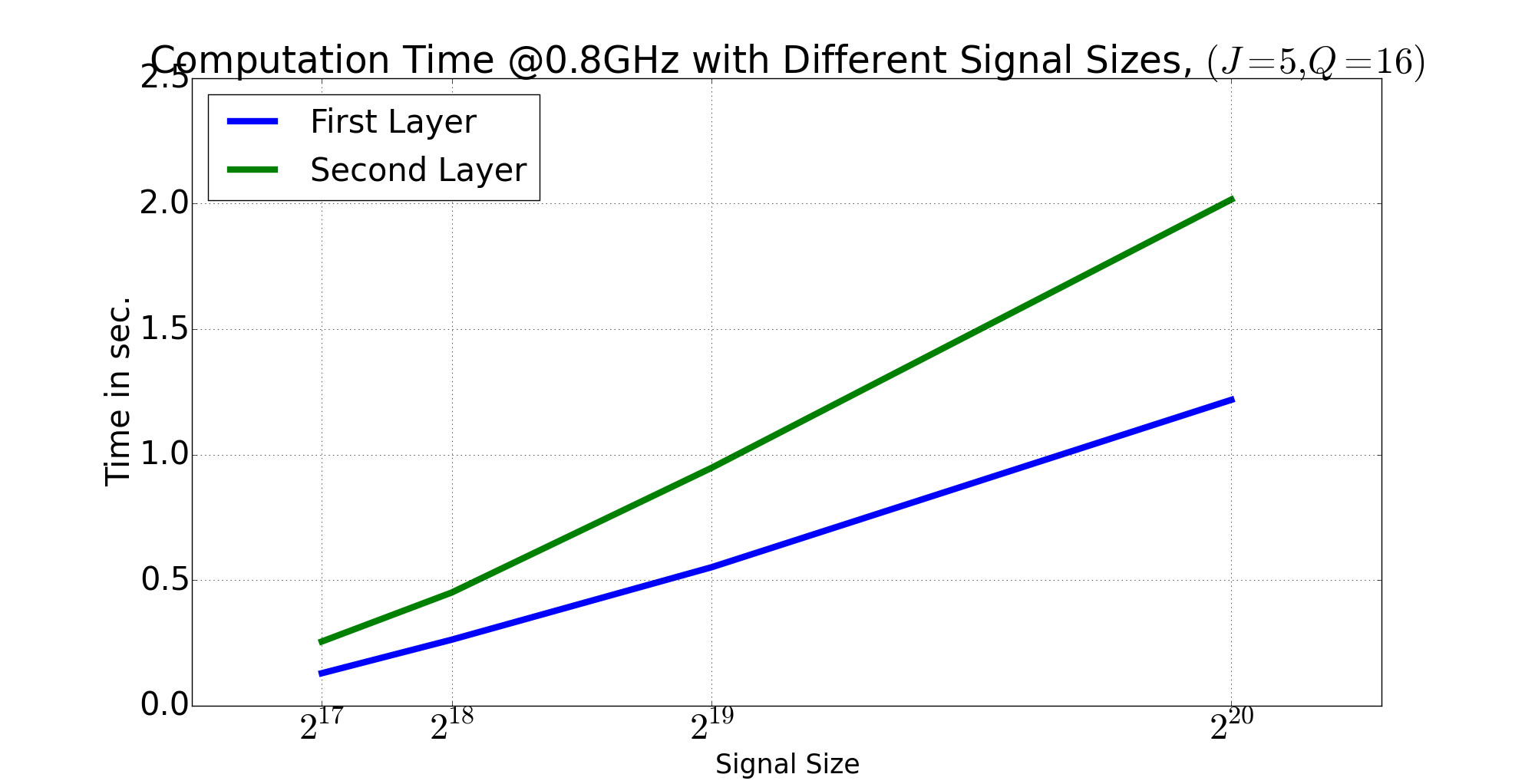}
\caption{Needed computation time to perform the transform for the first $\mathcal{X}^{(1)}[x]$ and second $\mathcal{X}^{(2)}[x]$ layers. The need computation time is more than $16$ times smaller than the known Scatnet toolbox implemented in C (\url{github.com/RandallBalestriero/CIGAL_GUI}) of $O(N \log (N))$ time complexity even when dealing with Fourier domain inputs. This shows the advantage of our approach which here is implemented in Python only.}
\label{fig::time}
\end{figure}

Finally, in Fig. \ref{fig::time} are presented some computational time for different input signals. We can see in this figure the high efficiency of our approach put in perspective of an existing C implementation of the scattering network. In fact, in this latter, one had to perform multiple inverse Fourier transforms in order to apply the nonlinearity and in order to compute the second layer for example apply again a Fourier transform and this for all the frequency bands. As a result the previously fastest known algorithm was of asymptotic complexity $O(N \log (N))$ even with a Fourier input. In addition, it did not leverage the sparsity of the representation leading to poor memory management and storing. Not however that for all the existing implementations, the complexity is linear with respect to the $J$ and $Q$ parameters.
Finally, with our framework, one can directly store the sparse matrices of the representations leading to space saving on the hard drive in addition of the actual Random Access Memory (RAM) saving during the computation.
\section{Validation on Bird Challenge}
\subsection{Dataset Presentation}
We now validate the use of the scattering coefficients as features for signal characterization through a supervised problem of audio recordings classification. The bird song classification challenge is made of $50$ classes and correspond to a small version of the BirdCLEF Challenge  \citep{joly2015lifeclef}. The recordings come from the Xeno-Canto database and mainly focus on species form South America. For our task, the dataset used for training is composed of  $924$ files sampled at $44100$ Hz. The valid set used to present our classification accuracy contains about $400$ files. Computing the $\mathcal{S}[x]$ and $\mathcal{V}[x]$ features on the training and valid set takes between $2$ to $3$ hours depending on the set of parameters used with a $2$-layer architecture on $1$ CPU. The files add up to a disk usage of $4.2$Go, the computed set of features however represent $450$Mo. As a result, we are able to encode and extract important characteristics of the signals while reducing the amount of redundant information.
We present in Fig.  \ref{fig::example1}\ref{fig::example2}\ref{fig::example3} examples of the dataset with the waveform as well as the representation $\mathcal{X}^{(1)}_{\lambda_1}[x]$. The aim is to first demonstrate the sparsity or high SNR in the physical domain involved by using a second order nonlinearity. In addition, one can see the different frequency modulated responses that could characterize a bird song. Overall, there are some fundamental difficulties in this task. The first challenge is to deal with translation invariance. In fact, the bird songs can be captured anywhere inside each files which themselves are of many different durations, from seconds to minutes. The second difficulty resides in characterizing well enough the time-frequency patterns of each specie without being altered by the ambient noise or possible presence of other species including human voices. Finally, difficulties also arise from the machine learning point of view with large class imbalance in the dataset.

\subsection{Results}
\begin{figure}[h]
\centering
\includegraphics[width=5.5in]{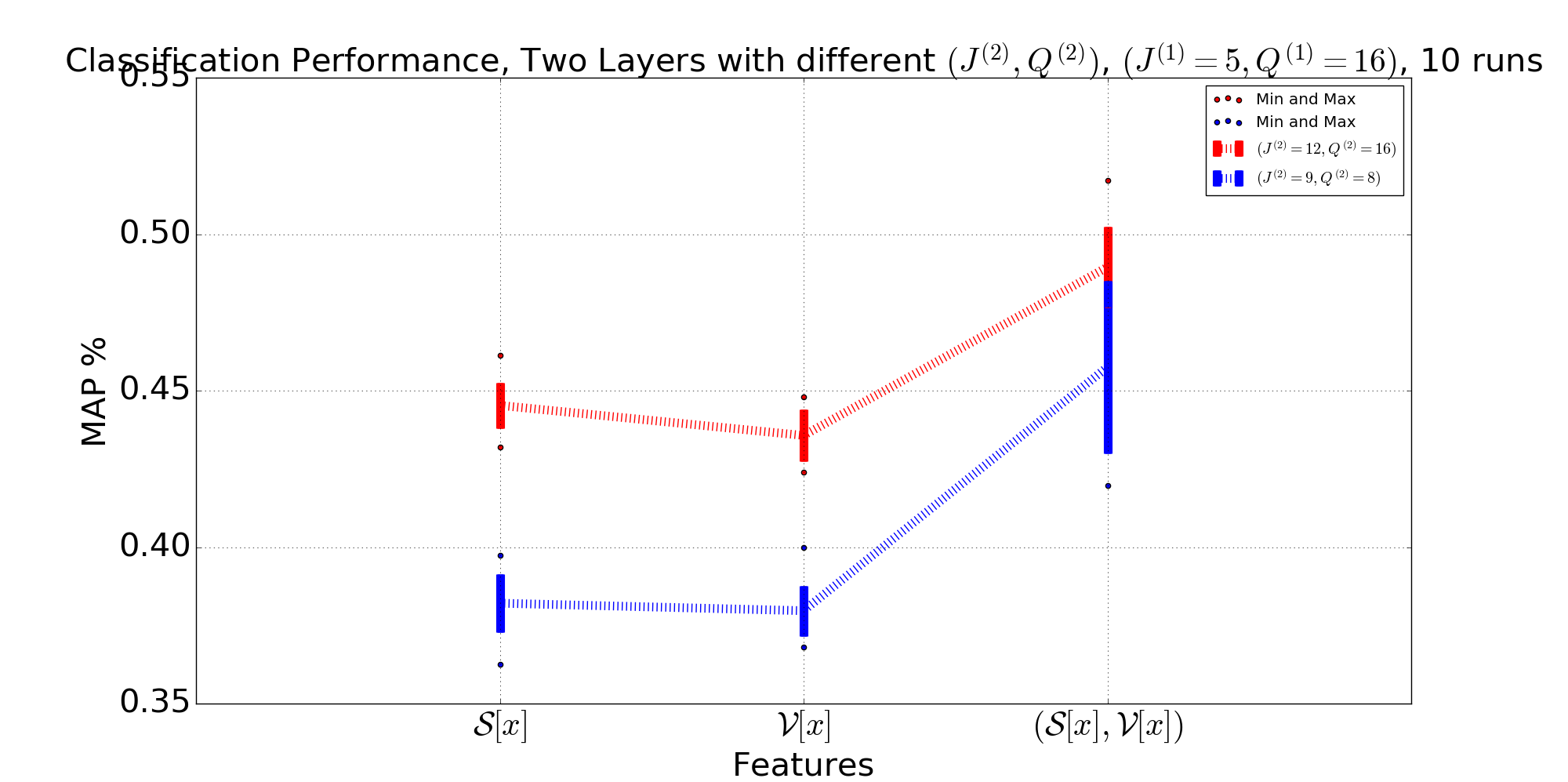}
\caption{Classification MAP given two set of parameters for the second scattering network layer. Are also presented the results when using only the scattering coefficients $\mathcal{S}[x]$, the dispersion coefficients $\mathcal{V}[x]$ and a concatenation of both, best is $52\%$.}
\label{fig::acc}
\end{figure}
We now present the classification results obtained via our developed framework. First of all, no additional features have been engineered and thus only the features developed in the paper are used. For the classification part, we decided to use a fast algorithm in accordance with the whole scheme developed earlier and thus used random forests  \citep{liaw2002classification}. In short, random forests are using bagging  of decision trees \citep{breiman1996bagging} and thus are able to aggregate multiple weak classifiers to end up with efficient class boundaries. One of its drawback resides in the fact that it can only create decision rule on each feature dimension without combining them as could do a logistic regression for example. In addition, we used a weighted loss function in order to deal with the imbalanced dataset  \citep{van2007experimental}.
Finally, no additional pre-processing/denoising has been used and no feature extraction/selection technique has been piped in. Yet, with this basic approach, we were able to reach an accuracy of $47.6\%$ and a Mean Average Precision (MAP) of $52.4\%$. The state-of-the-art technique for this problem reached a MAP of $53\%$  \citep{Challenge_data_ENS}. 
We present in Fig.\ref{fig::acc} some accuracy results where two sets of parameters have been used for the second layer of the scattering network. In addition, we show the classification results when using each features independently and combination of the two in order to highlight their complementarity. Given the deterministic transformation used and the lack of cross-validation and fine tuning, we think of these results as promising overall while being state-of-the-art if considering solutions where no learning was involved outside of the classifier.
For example, one extension on the classifier could be to use boosting algorithms \citep{schapire1998boosting} or neural networks. Concerning the representation, performing cross-validation on the parameters could lead to great improvements as finally a third scattering layered could also be considered.
\section{Conclusion}
We presented an extension of the scattering network in order to provide more discriminative time invariant features which are complementary to each others. The derivation of a second order invariant operator as well as the use of a second order nonlinearity in the layered representation computation led to efficient characterization of audio signals opening the door of many more possible time invariant features derivation. The whole framework has been derived in the Fourier domain in order to reach linear complexity in the input size as it is now possible to compute all the layer representations and feature without leaving the Fourier domain. Sparse storage is also a milestone of our algorithm leading to not only efficient computation but smart memory management allowing this framework to be applied online on energy efficient laptops and chips. In addition, only simple arithmetic operations are used and parallel implementation can be done easily as well as GPU portage. This framework can be applied without any assumption on the input signal and thus aims to be as general as possible as a unsupervised invariant feature extraction.
Finally, we hope to bring the consideration of sparse filters and Fourier based computation for deep convolutional networks. In fact, as the datasets get larger and larger, the complexity of the networks increase and convolutions might not be efficiently computed in the physical domain anymore. Since the convergence of the filter ensure their sparsity and smoothness, this consideration might help to bring deep learning to the family of scalable algorithms with the development of Fourier networks as a whole.
\section*{Acknowledgement}
We thank Institut Universitaire de France for the Glotin's Chair in 'Scene Analysis'. We thank SABIOD.ORG Mission Interdisciplinaire of the CNRS and Alexis Joly for co-organisation of the LifeClef Bird Challenge.
We also want to thank Mr. Romain Cosentino for his reviewing work and his help in bringing back in the physical domain some of our original sentences.

\bibliography{iclr2017_conference}
\bibliographystyle{iclr2017_conference}

\appendix
\section{Nonlinear Invariant in the Fourier domain}

\begin{align*}
||\mathcal{X}^{(l)}[x]-\mathcal{S}^{(l)}[x]||_2^2
=&\int \left( \mathcal{X}^{(l)}[x](t)-\mathcal{S}^{(l)}[x](t) \right)\\
&\left( \mathcal{X}^{(l)}[x](t)-\mathcal{S}^{(l)}[x](t) \right)^*dt\\
=&\int g(t)g(t)^*dt \;\;\text{ $g(t)=\mathcal{X}^{(l)}[x](t)-\mathcal{S}^{(l)}[x](t)$ }\\
=&\int \mathcal{F}[g](\omega)\mathcal{F}[g^*](\omega) d\omega \;\; \text{Plancherel Theorem}\\
=&||\mathcal{F}[g]||_2^2\\
=&||\mathcal{F}[\mathcal{X}^{(l)}[x]-\mathcal{S}^{(l)}[x]]||_2^2\\
=&||\mathcal{F}\left[ \mathcal{X}^{(l)}[x] \right]-\mathcal{F}\left[ \mathcal{S}^{(l)}[x] \right]||_2^2 \;\; \text{ Linear Operator}\\
=&||\mathcal{F}\left[ \mathcal{X}^{(l)}[x] \right]-\mathcal{F}\left[ \mathcal{X}^{(l)}[x]*\phi^{(l)} \right]||_2^2\\
=&||\mathcal{F}\left[ \mathcal{X}^{(l)}[x] \right]-\\
&\mathcal{F}\left[\mathcal{F}^{-1}\left[ \mathcal{F}[ \mathcal{X}^{(l)}[x]] \bigodot \mathcal{F}[ \phi^{(l)}] \right] \right]||_2^2\\
=&||\mathcal{F}\left[ \mathcal{X}^{(l)}[x] \right]- \mathcal{F}\left[ \mathcal{X}^{(l)}[x] \right] \bigodot \mathcal{F} \left[ \phi^{(l)} \right] ||_2^2\\
=&||\mathcal{F}\left[ \mathcal{X}^{(l)}[x] \right] \bigodot (1 - \mathcal{F} \left[ \phi^{(l)} \right]) ||_2^2.\\
\end{align*}

\section{Additional Material and Bird Song Representations}

\begin{figure}[h!]
\centering
\includegraphics[width=5.5in]{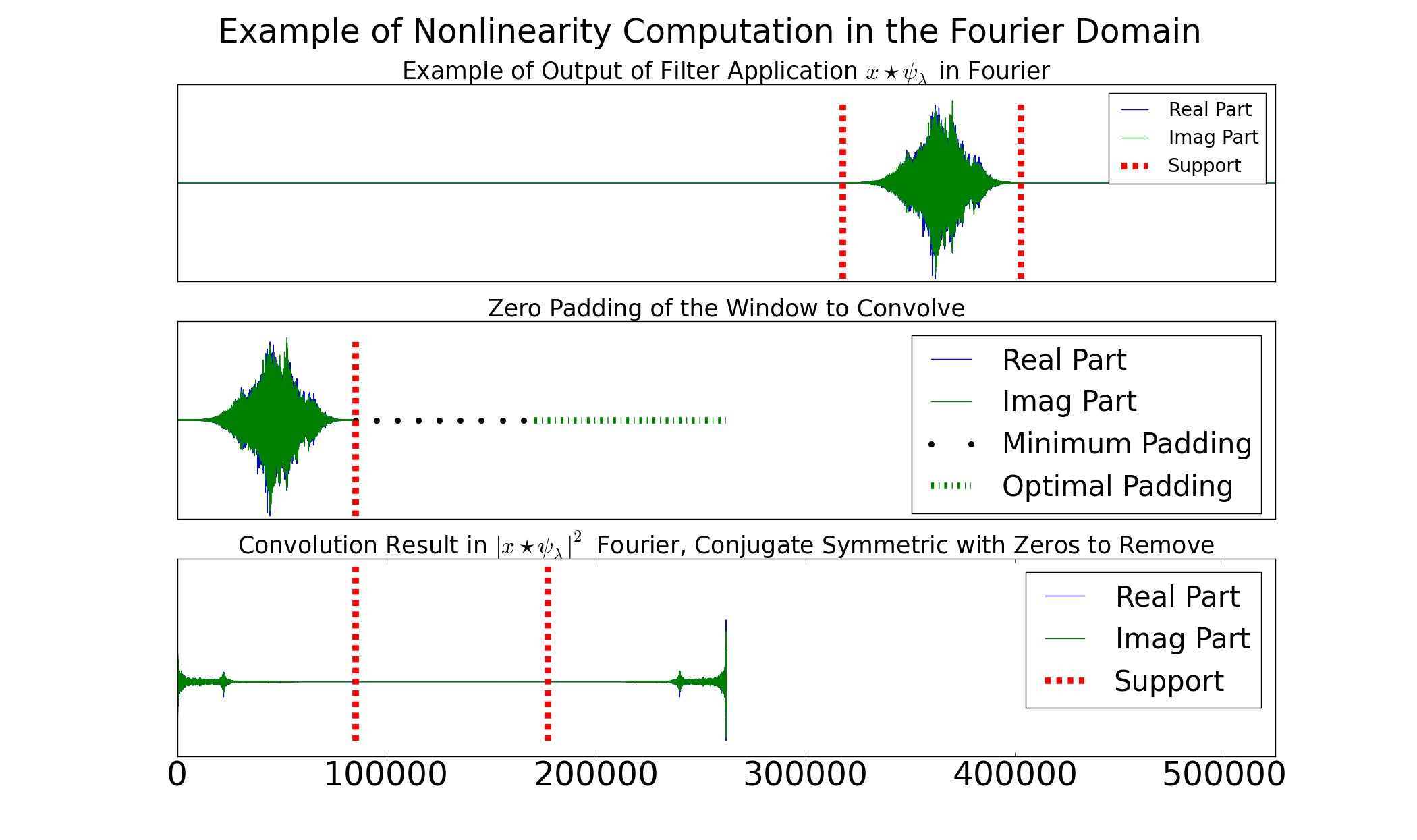}
\caption{Demonstration of the $\mathcal{P}$ computation in the Fourier domain. Top: input after application of a specific filter. Middle: Extracted window with nonzeros elements and optimal padding greater than the minimum size up to the next power of $2$. Bottom: Result of the convolution done through another Fourier transform and the convolution theorem, the kept coefficients are from $0$ to $M$ since they are followed by zeros and the complex conjugate of these coefficients leading to optimal results.}
\label{example_nonlinearity}
\end{figure}

\begin{figure}[h!]
\centering
\includegraphics[width=5.5in]{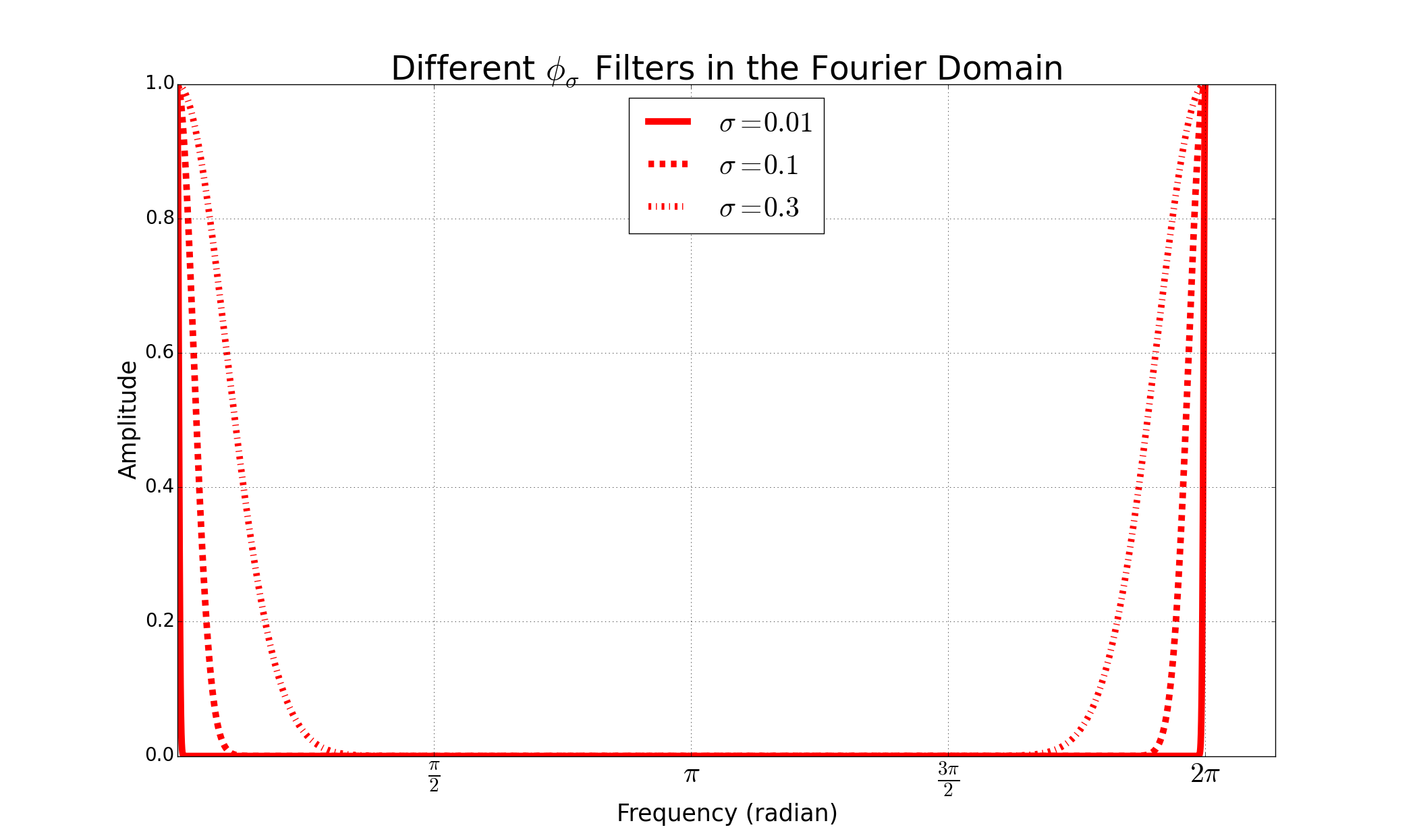}
\caption{Some possible $\phi$ filters in the Fourier domain corresponding to Gaussian filtering with bandwidth in the physical domain inversely proportional to the $\sigma$ in the Fourier domain without renormalization.}
\label{phis}
\end{figure}

\begin{figure}[h!]
\centering
\includegraphics[width=5.5in]{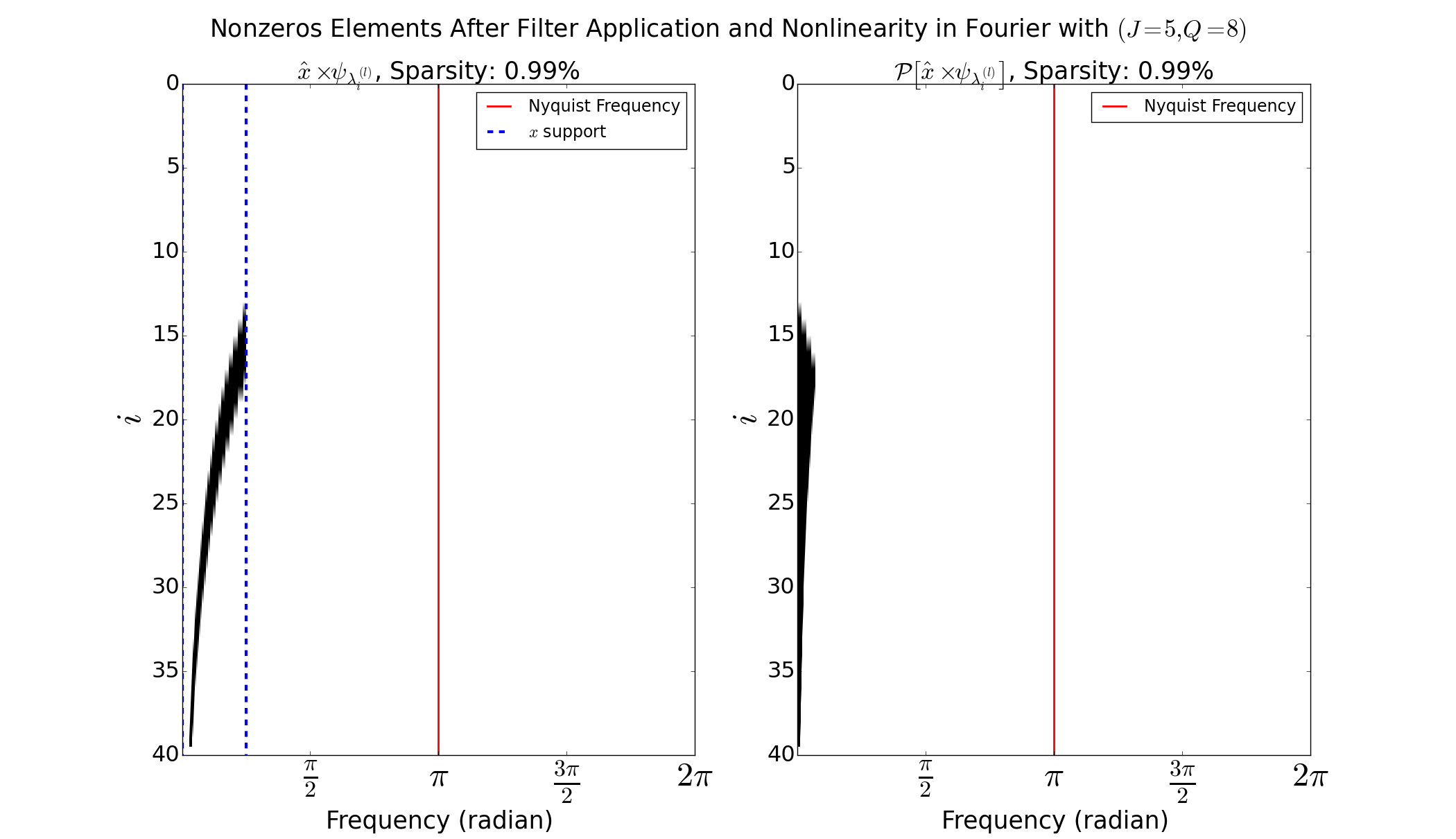}
\caption{Nonzeros elements present after application of the filters and the nonlinearity operator $\mathcal{P}$ on this representation for a sparse input.}
\label{fig::sparsity_matrix2}
\end{figure}

\begin{figure}[h!]
\centering
\includegraphics[width=5.5in]{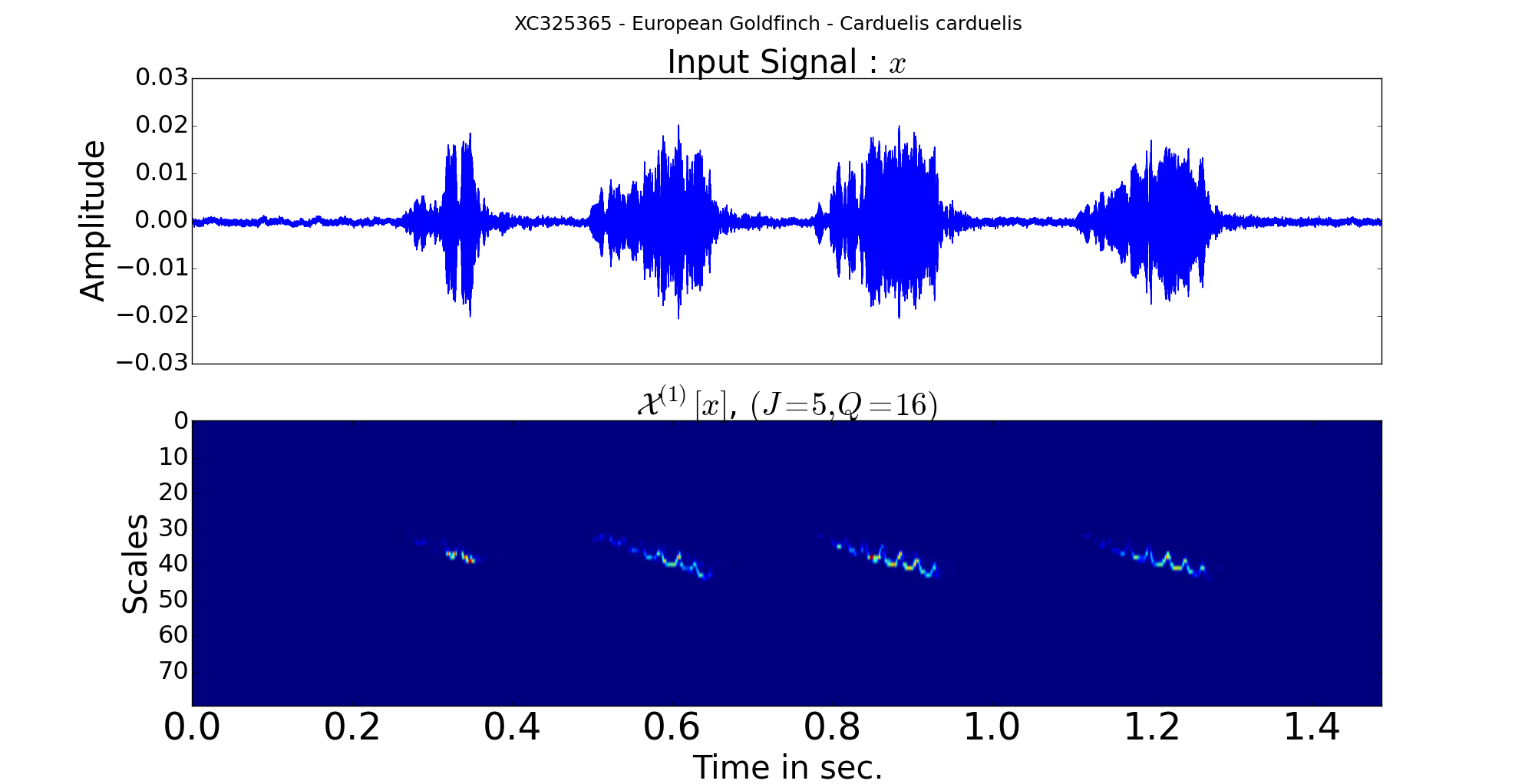}
\caption{Example 1: transform $\mathcal{X}^{(1)}_{\lambda_1}[x]$. In this case, clear frequency modulations appear for only one source and high SNR. The noise is contracted to $0$ through the nonlinearity.}
\label{fig::example1}
\end{figure}
\begin{figure}[h!]
\centering
\includegraphics[width=5.5in]{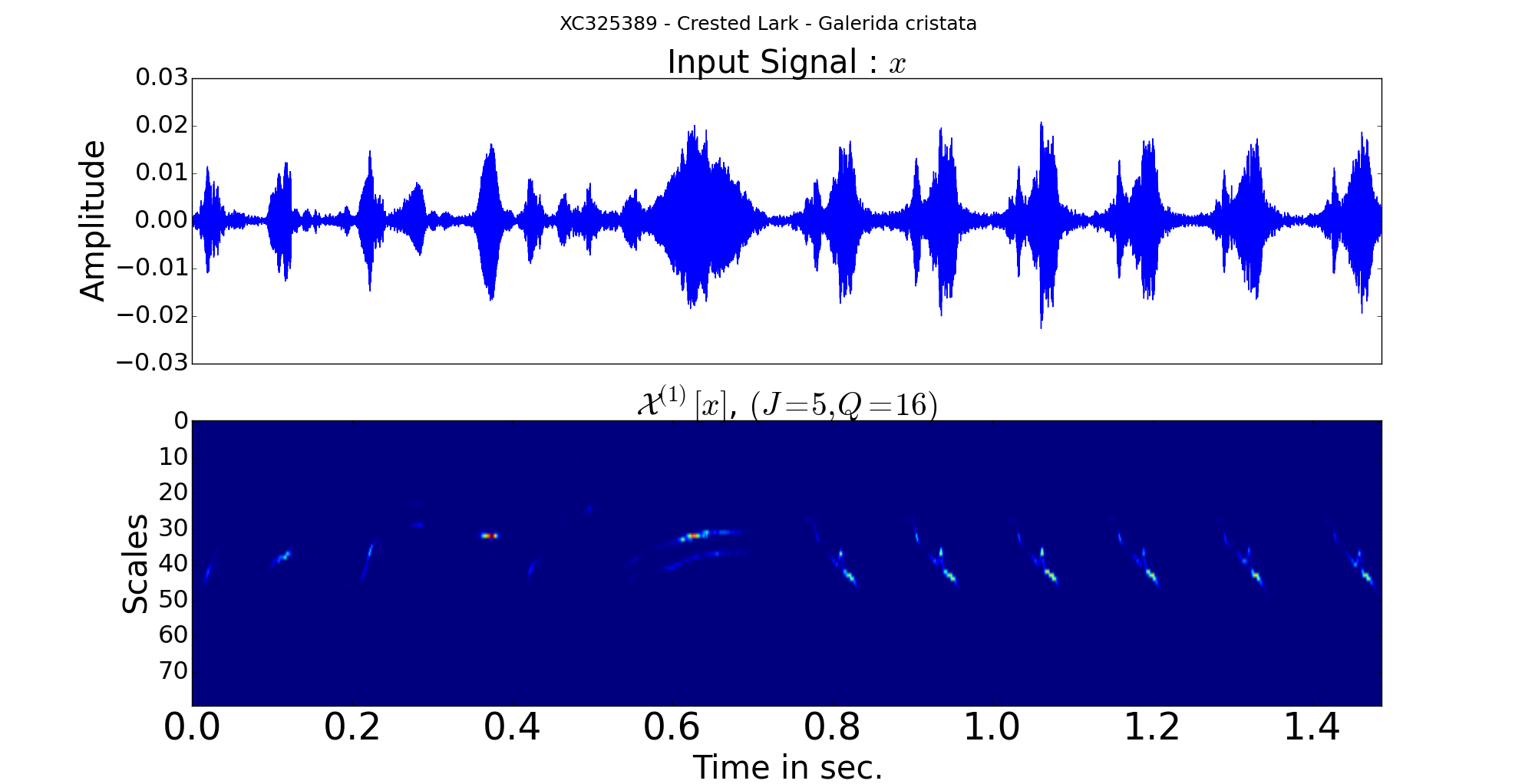}
\caption{Example 2: transform $\mathcal{X}^{(1)}_{\lambda_1}[x]$. In this case, the source presents multiple kinds of chirps, and frequency modulated patterns. Some harmonics are detected yet it is clear that aggregation of the time dimension with this representation only will aggregate the different patterns leading to poor signal characterization leading to the need of $\mathcal{X}^{(2)}_{\lambda_1,\lambda_2}[x]$.}
\label{fig::example2}
\end{figure}
\begin{figure}[h!]
\centering
\includegraphics[width=5.5in]{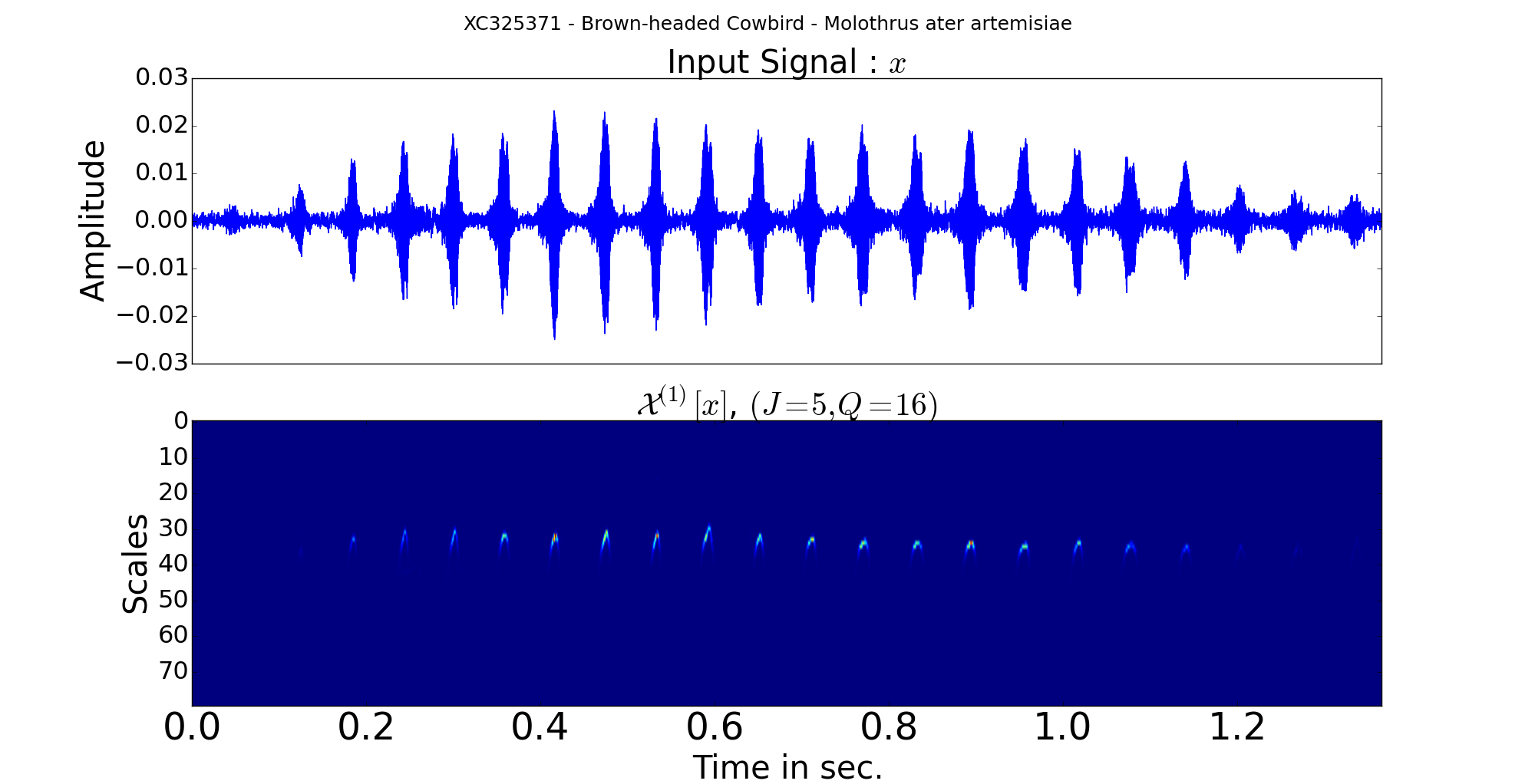}
\caption{Example 3: transform $\mathcal{X}^{(1)}_{\lambda_1}[x]$. In this case only transients are present and almost not frequency modulation appear on the features. This kind of signal will be well captured with one layer only.}
\label{fig::example3}
\end{figure}
\begin{figure}[h!]
\centering
\includegraphics[width=5.5in]{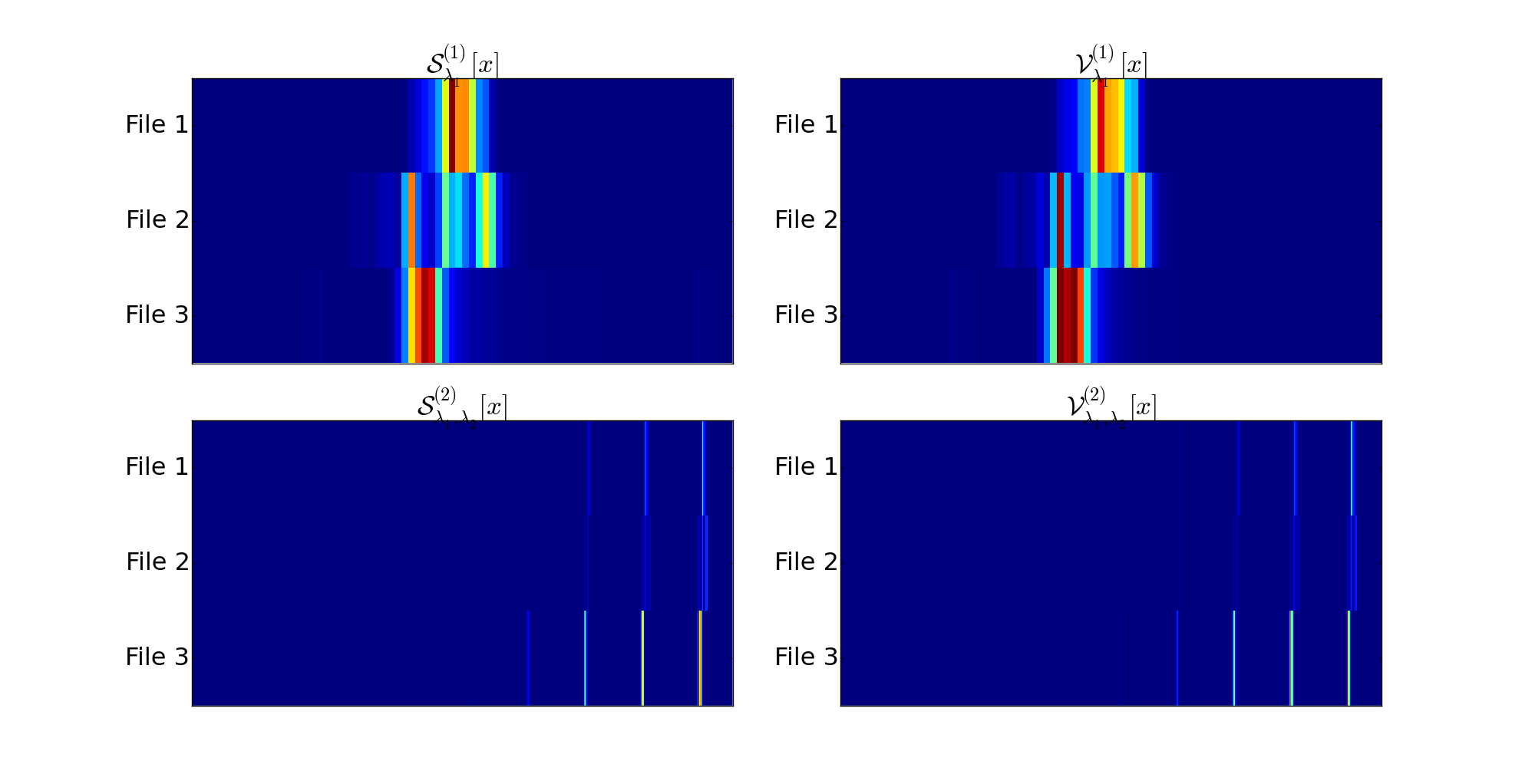}
\caption{We present here the features extracted form the $3$ examples presented in Fig. \ref{fig::example1},\ref{fig::example2},\ref{fig::example3}. The left part contains the scattering coefficients encoding the arithmetic mean whereas the right part concerns the dispersion coefficients. On the top part the features are extracted from the first layer and on the bottom are the features extracted form the second layer. It is clear that for these signals, the features of the first layer are enough to discriminate them. Notice that through global time invariance, one ends up with features vectors of exact same dimension for each signal and that they would remain the same if the input signal was translated.}
\label{fig::features}
\end{figure}

Using these three examples, we also present in Fig. \ref{fig::features} the resulting features computed on the first two layers of the scattering network in order to highlight the possibly linear hyperplanes separating these $3$ species in this new feature space.
\end{document}